%% file: main.tex
\documentclass[sigconf]{acmart}
\makeatletter
\global\@ACM@balancefalse
\makeatother
\usepackage{multirow}
\usepackage{enumitem}
\AtBeginDocument{%
  }

\copyrightyear{2026}
\acmYear{2026}
\setcopyright{cc}
\setcctype{by}
\acmConference[MM '26]{Proceedings of the 34th ACM International
  Conference on Multimedia}{November 10--14, 2026}{Rio de Janeiro, Brazil}
\acmBooktitle{Proceedings of the 34th ACM International Conference on
  Multimedia (MM '26), November 10--14, 2026, Rio de Janeiro, Brazil}
\acmISBN{979-8-4007-2213-4/2026/11}
\acmDOI{10.1145/3767308.3836401}
\begin{document}

\title{When Modalities Remember: Continual Learning for Multimodal Knowledge Graphs}

\author{Linyu Li}
\email{linyuli@stu.pku.edu.cn}
\affiliation{%
  \department{Key Laboratory of High Confidence Software Technologies (PKU),
    Ministry of Education}
  \institution{School of Computer Science, Peking University}
  \city{Beijing}
  \country{China}
}

\author{Zhi Jin}
\authornote{Corresponding author.}
\correspondingauthor
\email{zhijin@pku.edu.cn}
\affiliation{%
  \department{Key Laboratory of High Confidence Software Technologies (PKU),
    Ministry of Education}
  \institution{School of Computer Science, Peking University}
  \city{Beijing}
  \country{China}
}

\author{Yichi Zhang}
\affiliation{%
  \department{School of Computer Science}
  \institution{Zhejiang University}
  \city{Hangzhou}
  \country{China}
}

\author{Dongming Jin}
\author{Yuanpeng He}
\affiliation{%
  \department{Key Laboratory of High Confidence Software Technologies (PKU),
    Ministry of Education}
  \institution{School of Computer Science, Peking University}
  \city{Beijing}
  \country{China}
}

\author{Haoran Duan}
\affiliation{%
  \department{School of Cyber Science and Engineering}
  \institution{Wuhan University}
  \city{Wuhan}
  \country{China}
}

\author{Gadeng Luosang}
\affiliation{%
  \department{School of Information Science and Technology}
  \institution{Tibet University}
  \city{Lhasa}
  \country{China}
}

\author{Nyima Tashi}
\affiliation{%
  \department{School of Information Science and Technology}
  \institution{Tibet University}
  \city{Lhasa}
  \country{China}
}

\renewcommand{\shortauthors}{Li et al.}

\begin{abstract}
Real-world multimodal knowledge graphs (MMKGs) are dynamic, with new entities, relations, and multimodal knowledge emerging over time. Existing continual knowledge graph reasoning (CKGR) methods focus on structural triples and cannot fully exploit multimodal signals from new entities. Existing multimodal knowledge graph reasoning (MMKGR) methods, however, usually assume static graphs and suffer catastrophic forgetting as graphs evolve. To address this gap, we present a systematic study of continual multimodal knowledge graph reasoning (CMMKGR). We construct several continual multimodal knowledge graph benchmarks from existing MMKG datasets and propose MRCKG, a new CMMKGR model. Specifically, MRCKG employs a multimodal-structural collaborative curriculum to schedule progressive learning based on the structural connectivity of new triples to the historical graph and their multimodal compatibility. It also introduces a cross-modal knowledge preservation mechanism to mitigate forgetting through entity representation stability, relational semantic consistency, and modality anchoring. In addition, a multimodal contrastive replay scheme with a two-stage optimization strategy reinforces learned knowledge via multimodal importance sampling and representation alignment. Experiments on multiple datasets show that MRCKG preserves previously learned multimodal knowledge while substantially improving the learning of new knowledge.
\end{abstract}

\ccsdesc[500]{Computing methodologies~Knowledge representation and reasoning}

\keywords{Knowledge Graph, Continual Learning, Multimodal Learning, Multi-modal knowledge graph reasoning}

\maketitle

\section{Introduction}

Multimodal Knowledge Graphs (MMKGs)\cite{zhu2022multi,liang2024survey,chen2024knowledge,wang2023tiva} introduce multimodal information such as images and text onto the traditional triplet structure, providing richer and more discriminative entity representations. As an important knowledge foundation in the Multimedia Reasoning\cite{li2025entity,wang2025medkcoop,cao2022cross,wu2024mkg}, MMKGs typically learn entity and relation representations using Multimodal knowledge graph reasoning (MMKGR) \cite{zhao2025dark,cao2022otkge,fang2025cdib,liang2024simple,zhao2024contrast,zhang2025tokenization,li2025towards}
for various downstream tasks. However, real-world knowledge graphs are constantly evolving, with new entities, relations, and facts constantly emerging\cite{li2026learning,liu2024towards,zhao2025rethinking,yang2025knowledge}. Retraining the model from scratch for each update is not only costly but also fails to meet timeliness requirements. Furthermore, most existing MMKGR models are based on static graph assumptions, making them ill-suited to the ever-increasing volume of entities, relations, and multimodal information in real-world scenarios. This raises a question that has not yet been systematically studied: how can we absorb new knowledge while avoiding forgetting old knowledge without full retraining when MMKGs are constantly evolving? This problem is even more challenging than traditional Continual Knowledge Graph Reasoning (CKGR). Based on this, we constructed multiple continual benchmark datasets using the existing MMKG dataset through three different evolutionary approaches, and systematically proposed the Continual Multimodal Knowledge Graph Reasoning (CMMKGR) task.
\begin{figure}[t]
    \centering
    \includegraphics[width=\linewidth]{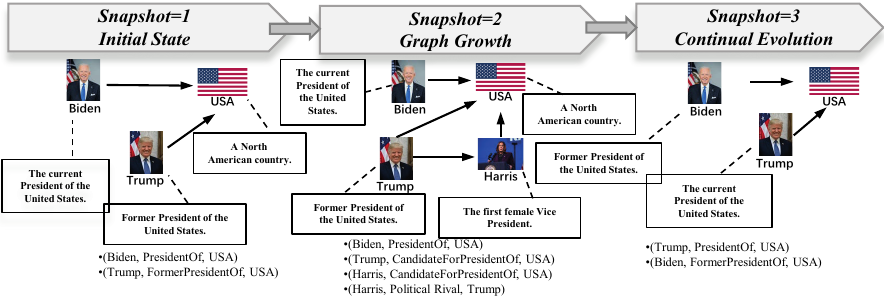}
    \caption{CMMKG stores knowledge in the form of triplets; however, unlike traditional knowledge graphs, its multimodal information also continuously evolves over time.}
    \Description{Three snapshots illustrate a continually evolving multimodal knowledge graph. Entities representing political figures and the United States gain and change structural relations, portrait images, and textual descriptions over time.}
    \label{fig:cmmkg-evolution}
\end{figure}
Taking Figure \ref{fig:cmmkg-evolution} as an example, at a given snapshot, the graph already contains structured triples about national leaders, along with official portraits, news photos, and biographical text. This creates a trade-off: retraining the entire graph from scratch is too costly, while continual learning based only on structural information overlooks key multimodal signals, such as headshots, campaign posters, and news text, which are essential for distinguishing political figures and tracking changes in their relationships.
We summarize three main challenges facing CMMKGR. First, new entities are often weak in structure but strong in multimodal signals, so the model has to rely on images and text to handle the cold-start stage. Second, in continual learning, what gets disrupted is not just structural embeddings, but also visual projections, textual projections, and cross-modal alignment, which can weaken the multimodal semantics of previously learned knowledge. Third, the order of learning new knowledge affects incremental performance. Samples that are closer to old entities in graph structure or in visual and textual semantics are usually easier to learn, so they are better starting points for training.

Based on the above analysis, we argue that in continual learning, multimodal information is not merely auxiliary evidence for improving embedding quality, but also serves as a stable semantic anchor for mitigating forgetting. Compared with evolving structural representations, pretrained visual and textual features remain relatively stable over snapshots and can therefore provide reliable anchors for continuously updated representations. To this end, we propose MRCKG, a model tailored to the CMMKGR task. Specifically, we first introduce Multimodal Structure-aware Curriculum Learning (MSCL), which jointly evaluates the learning priority of new triples using structural connectivity and multimodal compatibility, allowing the model to absorb new knowledge progressively from easy to hard. We then design Cross-Modality Knowledge Preservation (CMKP), which unifies entity stability, consistency of relational semantic patterns, and modal anchoring within a single preservation objective. Finally, we develop Multimodal Contrastive Replay, or MMCR, together with a two-stage optimization strategy, to further consolidate previously learned knowledge through multimodal importance sampling and contrastive replay.

In comprehensive comparisons with multiple baselines, MRCKG achieves consistently higher MRR scores than existing methods across several datasets. The experiments also show that simple multimodal fusion is almost ineffective in continual learning, whereas the dedicated design of MRCKG enables multimodal information to genuinely function as a semantic anchor. The main contributions of this work are summarized as follows:
\par (1) We systematically formulate the CMMKGR task, summarize its core challenges, including multimodal catastrophic forgetting and cross-modal consistency preservation, and construct nine benchmark datasets for CMMKGR.
\par (2) We propose MRCKG, a CMMKGR model with three key modules: MSCL, CMKP, and MMCR. Specifically, MSCL progressively absorbs new knowledge, CMKP preserves the structural and semantic memory of previously learned knowledge in a unified manner, and MMCR further consolidates historical knowledge through multimodal contrastive replay.
\par (3) Extensive experiments on multiple benchmark datasets reveal an important finding: multimodal information can serve as a semantic anchor in continual knowledge graph reasoning.

\section{Related Work}

Recent KGC research has enhanced static completion through multi-view Riemannian manifold fusion\cite{li2025multi} and alignment--distillation-based multilingual data augmentation\cite{li2026a2da}. These geometric and linguistic advances do not address knowledge retention under continual graph evolution.

\subsection{Multimodal Knowledge Graph Reasoning}

MMKGR explicitly incorporates visual\cite{wang2021visual}  and textual information into entity representation learning to make up for the limits of purely structural methods\cite{chen2024knowledge}. In multimodal fusion, MKGformer\cite{chen2022hybrid}, IMF\cite{li2023imf}, and LAFA\cite{shang2024lafa} explore fusion strategies from the perspectives of cross-modal Transformers, interaction mechanisms, and neighbor structure information, respectively. More recent studies further improve fusion granularity and robustness, including the fine-grained tokenization of MYGO\cite{zhang2025tokenization}, the attention penalty of APKGC\cite{jian2025apkgc}, the frequency-domain fusion of WFF\cite{xu2025wff}, the dynamic structure awareness of DySaRL\cite{liu2024dysarl}, the structure-aware multimodal modeling of Li et al.\cite{li2025towards}, the segmentation-based similarity enhancement of SSEF\cite{wang2025segmentation}, and the tokenization-decoupling strategy of TFD\cite{su2026tokenization}.

Another line of work focuses on modality separation and cross-modal interaction. MoSE\cite{zhao2022mose} learns separate relation representations for each modality and combines their decisions. NativE\cite{zhang2024native} introduces a relation-guided dual adaptive fusion scheme, CDIB\cite{fang2025cdib} uses the information bottleneck to model cross-modal consistency, and RMD\cite{zhao2025dark} explores complementary relations across modalities through reinforced distillation. At the training level, OTKGE\cite{cao2022otkge} performs cross-modal alignment with optimal transport, MMRNS\cite{xu2022relation} and DHNS\cite{niu2025diffusion} generate high-quality negative samples through relation enhancement and diffusion models, VISTA\cite{lee2023vista}, SimDiff\cite{li2024simdiff} and SatMKGR\cite{li2025meta} augment data by synthesizing triplets, and CMR\cite{zhao2024contrast} addresses inductive completion by combining contrastive learning with semantic neighbor retrieval.

In addition, MMKGR has been extended to few-shot completion\cite{wei2024multi}, cross-graph entity alignment\cite{ni2023psnea,wang2025explicit}, multimodal entity linking\cite{kim2025kgmel,luo2024bridging}, and MMKG-based reasoning enhancement for large models\cite{lee2024multimodal}. Although these studies clearly show the value of multimodal information for KGR, they all rely on the assumption of static graphs and therefore cannot directly address catastrophic forgetting and cross-modal semantic drift caused by the continuous evolution of knowledge graphs.

\subsection{Continual Learning for Knowledge Graph Reasoning}
Continual knowledge graph reasoning (CKGR) aims to enable models to acquire new knowledge while avoiding catastrophic forgetting\cite{mccloskey1989catastrophic}. Existing methods mainly follow three lines. First, regularization-based methods preserve old knowledge by constraining updates to important parameters. Typical examples include EWC\cite{kirkpatrick2017overcoming}, which is based on the Fisher information matrix, and SI\cite{zenke2017continual}, which relies on online contribution tracking. Second, architecture-based and replay-based methods mitigate forgetting through structural expansion or data replay. Representative studies include PNN\cite{rusu2016progressive}, EMR\cite{wang2019sentence}, and DiCGRL\cite{kou2020disentangle}. Third, distillation-based and adapter-based methods seek a balance between efficiency and knowledge retention. Examples include the incremental distillation strategy of IncDE\cite{liu2024towards}, the adaptive low-rank adapters of FastKGE\cite{liu2024fast}, and the forgetting-mitigating modulation of MoFot\cite{jiang2026towards}. Recent studies have further advanced these directions. Bayesian-guided continual embedding guides knowledge graph evolution across snapshots\cite{li2026learning}. LKGE\cite{cui2023lifelong} combines masked autoencoders with transfer regularization. SAGE\cite{li2025sage} and ERPP\cite{yang2025knowledge} further improve continual learning from the perspectives of adaptive dimensional expansion and relational path propagation, respectively. CFKGC\cite{li2024learning} extends CKGE to the few-shot setting. However, all these methods are designed for unimodal structured knowledge graphs. They do not consider the joint evolution of multimodal information and therefore cannot directly address cross-modal semantic drift in continual learning for multimodal knowledge graphs.

\section{Preliminaries and Problem Definition} 

\textbf{Definition 1 (Multimodal Knowledge Graph).} A multimodal knowledge graph is defined as a quintuple $\mathcal{G} = (\mathcal{E}, \mathcal{R}, \mathcal{T}, \mathcal{V}, \mathcal{D})$, where $\mathcal{E}$ denotes the entity set, $\mathcal{R}$ denotes the relation set, and $\mathcal{T} \subseteq \mathcal{E} \times \mathcal{R} \times \mathcal{E}$ denotes the set of triples. Each entity $e \in \mathcal{E}$ is associated with a set of images $\mathcal{V}(e) = \{v_e^1, \ldots, v_e^{k_v}\}$ and a textual description $\mathcal{D}(e) = \{w_e^1, \ldots, w_e^{k_d}\}$.

\textbf{Definition 2 (Evolving Snapshot Sequence).} Given an ordered set of snapshot indices $i = 0, 1, \ldots, T{-}1$, a multimodal knowledge graph forms an evolving snapshot sequence $\{\mathcal{S}_0, \mathcal{S}_1, \ldots, \mathcal{S}_{T-1}\}$, where each snapshot $\mathcal{S}_i = (\mathcal{E}_i, \mathcal{R}_i, \mathcal{T}_i, \mathcal{V}_i, \mathcal{D}_i)$. The sequence satisfies the monotonic expansion property: $\mathcal{E}_{i-1} \subseteq \mathcal{E}_i$, $\mathcal{R}_{i-1} \subseteq \mathcal{R}_i$, and $\mathcal{T}_{i-1} \subseteq \mathcal{T}_i$. The newly added triples and entities are defined as $\Delta\mathcal{T}_i = \mathcal{T}_i \setminus \mathcal{T}_{i-1}$ and $\Delta\mathcal{E}_i = \mathcal{E}_i \setminus \mathcal{E}_{i-1}$, respectively ($i \geq 1$; when $i = 0$, $\Delta\mathcal{T}_0 = \mathcal{T}_0$ and $\Delta\mathcal{E}_0 = \mathcal{E}_0$).

\textbf{Definition 3 (CMMKGR Task).}
Given a scoring function $f_\theta: \mathcal{E} \times \mathcal{R} \times \mathcal{E} \to \mathbb{R}$ parameterized by $\theta$, link prediction aims to rank the correct answer highest among all candidate entities for a query $(h,r,?)$ or $(?,r,t)$. In the continual multimodal knowledge graph reasoning (CMMKGR) setting, when learning on snapshot $\mathcal{G}_i$, the model updates its parameters from $\theta_{i-1}$ to $\theta_i$ using the newly arrived triples $\Delta\mathcal{T}_i$, the associated multimodal information of newly introduced entities $\Delta\mathcal{E}_i$, and an optional bounded replay memory $\mathcal{M}_{i-1}$ that stores historical samples from previous snapshots, rather than being retrained on the full graph $\mathcal{T}_i$. After training on $\mathcal{G}_i$, the model is evaluated on the test sets of all observed snapshots, i.e.,$\bigcup_{j=0}^{i} \mathcal{G}^{\mathrm{test}}_j$.
The goal is to maximize link prediction performance on both newly acquired and previously learned knowledge, measured by MRR and Hits@$K$, while achieving a favorable trade-off between plasticity and stability.

\section{Method}
\subsection{Framework Overview}
\begin{figure*}[t]
    \centering
    \includegraphics[width=0.9\linewidth]{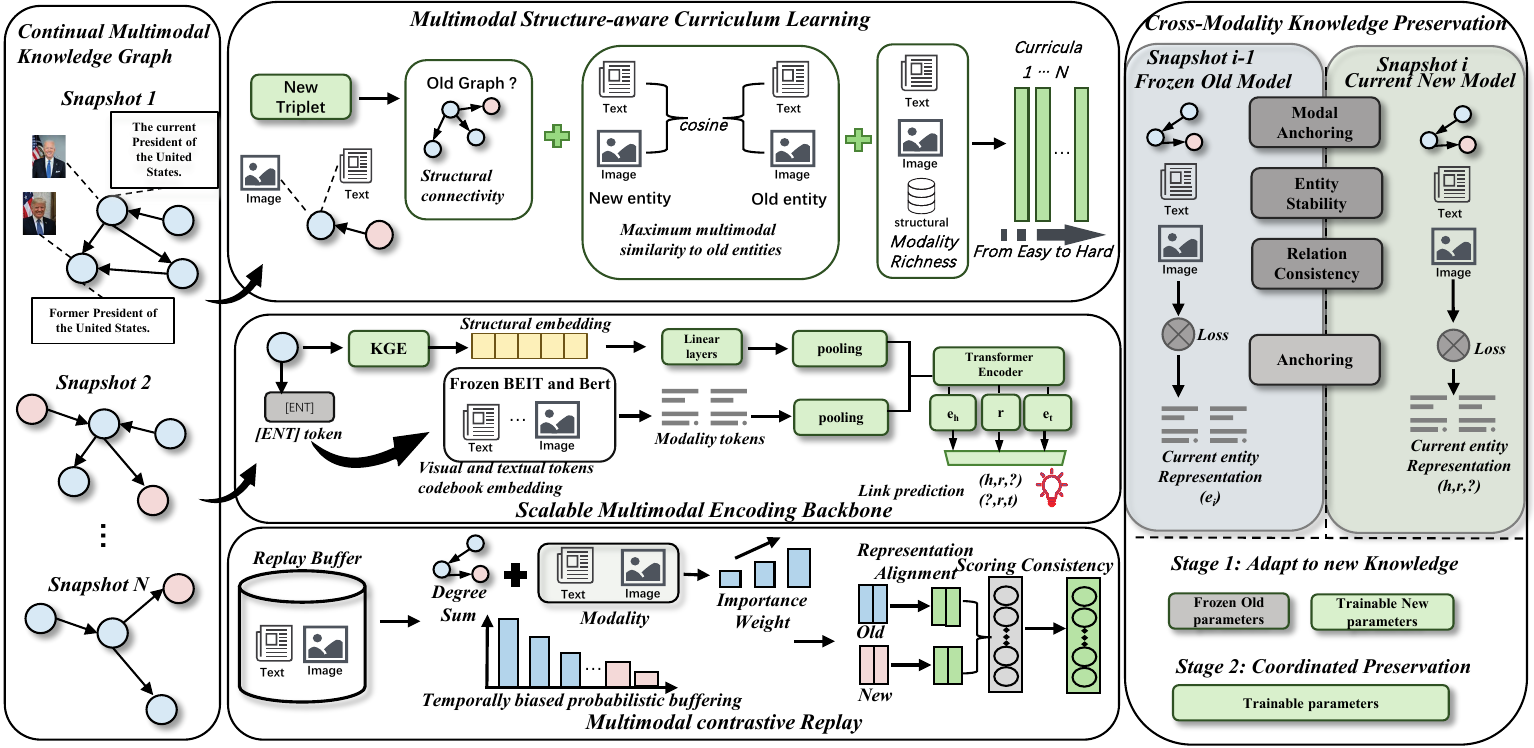}
    \caption{Overall framework of MRCKG for continual multimodal knowledge graph reasoning.}
    \Description{The MRCKG pipeline combines a scalable multimodal encoding backbone with multimodal structure-aware curriculum learning, cross-modality knowledge preservation, multimodal contrastive replay, and two-stage optimization across graph snapshots.}
    \label{fig2}
\end{figure*}
We propose MRCKG, a unified framework for continual multimodal knowledge graph embedding. Its design is motivated by three key observations: (1) the order in which new knowledge arrives directly affects training stability, and relying solely on graph-structural ordering is insufficient for multimodal settings, where newly introduced entities may be semantically similar yet structurally isolated; (2) catastrophic forgetting is not limited to structural entity embeddings, but also appears in the visual and textual projection layers, relational semantic patterns, and cross-modal alignment as parameters are updated; and (3) frozen pretrained visual and textual features are naturally stable across snapshots and can therefore serve as semantic anchors in continual learning.

Building on these insights, MRCKG consists of a scalable multimodal encoding backbone and three collaborative mechanisms: MSCL, which jointly leverages structural connectivity and multimodal semantic similarity to progressively rank new samples and determine the order in which new knowledge is absorbed; CMKP which constrains the drift of previously learned knowledge from three perspectives, namely entity stability, consistency of relational semantic patterns, and modal anchoring; and MMCR, which selects representative historical samples through multimodal-aware importance sampling and reinforces past knowledge via contrastive-consistent replay. These three components work together around the core idea of multimodal semantic anchors.

At snapshot $\mathcal{S}_i$, the input consists of the newly added triplet set $\Delta T_i$ together with the images and texts associated with the entities involved. MSCL first performs progressive ranking of the training samples; the encoding backbone then learns representations for entities and relations; CMKP constrains the semantic drift of old knowledge in the structural, relational, and modal spaces; and MMCR samples historical instances from the replay buffer and mixes them with the current batch for joint training. The entire training process follows a two-stage optimization strategy: new entities are first adapted, and then a global coordination step is performed.

\subsection{Scalable Multimodal Encoding Backbone}

We treat the encoder as the foundational backbone of the whole method rather than as an independent contribution. For each entity $e$, we construct the following input sequence and concatenate it before feeding it into the encoder:
\begin{equation}\label{eq:input}
\mathbf{X}_{in}(e) = [\mathrm{ENT}] \oplus \mathbf{s}_e \oplus \hat{\mathbf{v}}_{e,1:m} \oplus \hat{\mathbf{w}}_{e,1:n},
\end{equation}
where $\oplus$ denotes sequence concatenation, $[\mathrm{ENT}]$ is a global aggregation token, $\mathbf{s}_e \in \mathbb{R}^d$ is the learnable structural embedding of entity $e$, and $\hat{\mathbf{v}}_{e,1:m}$ and $\hat{\mathbf{w}}_{e,1:n}$ are the visual and textual token representations after projection through learnable linear layers, respectively. Structural embeddings are obtained by training KGE models.

On the visual side, we use a frozen BEiT~\cite{bao2021beit} visual tokenizer and codebook embedding layer to extract discrete visual token representations. On the textual side, we use a frozen BERT~\cite{devlin2018bert} tokenizer and word embedding layer to obtain textual token representations. These are then mapped into the same $d$-dimensional space through two learnable linear projection layers, $\mathbf{W}_v \in \mathbb{R}^{d_v \times d}$ and $\mathbf{W}_w \in \mathbb{R}^{d_w \times d}$. To facilitate the definition of the subsequent loss functions, we apply mean pooling to the projected token sequences and obtain the modality-level aggregated representations for each entity:
\begin{equation}\label{eq:pool}
\bar{\mathbf{v}}_e = \frac{1}{m}\sum_{j=1}^{m}\hat{\mathbf{v}}_{e,j},\qquad
\bar{\mathbf{w}}_e = \frac{1}{n}\sum_{j=1}^{n}\hat{\mathbf{w}}_{e,j}.
\end{equation}
If entity $e$ lacks one modality, the corresponding $\bar{\mathbf{v}}_e$ or $\bar{\mathbf{w}}_e$ is set to a zero vector, and this entity is skipped in the relevant loss terms for the missing modality. After the sequence above is fed into the Transformer encoder, we take the output at the $[\mathrm{ENT}]$ position as the entity representation $\mathbf{e}$. Relations are still represented by learnable embeddings $\mathbf{r}$. To strengthen contextual interaction within triplets, we further feed the head entity, relation, and tail entity into a Contextual Encoder,  and perform link prediction using the scoring function of the KGE model.

When new entities $\Delta\mathcal{E}_i$ arrive, we only need to allocate new structural embeddings $\mathbf{s}_e$ for them and attach the corresponding visual and textual token indices, without rebuilding the entire multimodal encoder. In this way, the model can both inherit the multimodal representation capability already learned and naturally adapt to the continual expansion of the entity set.

\subsection{Multimodal Structure-aware Curriculum Learning (MSCL)}

\subsubsection{Curriculum Score}

Let the set of historical entities at the beginning of the current snapshot be $\mathcal{E}_{old}=\mathcal{E}_{i-1}$. For any newly arrived triplet $(h,r,t)\in \Delta\mathcal{T}_i$, we define its curriculum score as
\begin{equation}\label{eq:phi}
\phi(h,r,t)=\alpha \cdot c_{str}(h,t)+\beta \cdot c_{mm}(h,t\mid \mathcal{E}_{old})+\gamma \cdot c_{rich}(h,t),
\end{equation}
where $\alpha$, $\beta$, and $\gamma$ are weighting coefficients. The three terms measure structural connectivity, multimodal compatibility, and modality richness, respectively. The structural term is defined as $c_{str}(h,t)=\mathbb{I}[h\in \mathcal{E}_{old}\ \text{or}\ t\in \mathcal{E}_{old}]$, a binary indicator that equals 1 if either the head or the tail entity has already appeared in the old graph, suggesting that the triplet can be more easily absorbed by the current model. We define $c_{rich}(h,t)=\bigl(M(h)+M(t)\bigr)/2$, where $M(e)$ denotes the modality richness of entity $e$ (for example, whether it has both image and text modalities, and whether the number of tokens is sufficient), normalized to $[0,1]$.

The multimodal compatibility term $c_{mm}$ measures the maximum semantic similarity between new entities and old entities in the frozen pretrained feature space. Let $\mathcal{U}_{new}(h,t)=\{u\in\{h,t\}\mid u\notin \mathcal{E}_{old}\}$ denote the set of new endpoints in the triplet. We define the multimodal similarity between a new entity and an old entity as
\begin{equation}\label{eq:sim_mm}
\mathrm{sim}_{mm}(u,e')=\eta_v\cdot \cos(\mathbf{v}_u^{pt},\mathbf{v}_{e'}^{pt})+\eta_t\cdot \cos(\mathbf{w}_u^{pt},\mathbf{w}_{e'}^{pt}),
\end{equation}

where $\mathbf{v}_e^{pt}$ and $\mathbf{w}_e^{pt}$ are the raw outputs of the frozen pretrained encoders, and $\eta_v$ and $\eta_t$ are modality balancing coefficients. When $\mathcal{U}_{new}\neq\varnothing$, we set
\[
c_{mm}(h,t \mid \mathcal{E}_{old})
=
\max_{u\in\mathcal{U}_{new}}
\max_{e'\in \mathcal{E}_{old}}
\mathrm{sim}_{mm}(u,e').
\]
Otherwise, $c_{mm}=0$.

\subsubsection{Progressive Training Procedure}

After sorting $\Delta\mathcal{T}_i$ in descending order according to $\phi(h,r,t)$, we divide it into $K$ curricula and train them one by one in sequence. Once each curriculum is completed, the new entities it contains are added to the known entity set, and the $c_{str}$ term for the remaining curricula is recalculated, while $c_{mm}$ and $c_{rich}$ stay unchanged. In this way, the model first encounters new samples that are more connected to the old graph and more similar to previously learned knowledge, and then gradually moves toward more isolated subgraph regions, using multimodal information to support the cold start of new entities.

\subsection{Cross-Modality Knowledge Preservation (CMKP)}

\subsubsection{Entity-Level Stability}

For each historical entity $e\in \mathcal{E}_{i-1}$, we first constrain its full representation to remain stable across snapshots:
\begin{equation}\label{eq:str}
\mathcal{L}_{str}=\sum_{e\in \mathcal{E}_{i-1}}\lambda_e\cdot \mathcal{D}\bigl(\mathbf{e}^{(i)},\mathbf{e}^{(i-1)}\bigr),
\end{equation}
where $\mathcal{D}(\cdot,\cdot)$ denotes the Smooth L1 loss. When its inputs are vectors, $\mathcal{D}$ is applied element-wise and aggregated into a scalar penalty; when its inputs are scalars, it reduces to the standard scalar Smooth L1 loss. Here $\lambda_e$ is the importance weight of entity $e$.
Rather than treating all entities equally, we define this weight according to its role in the graph structure and the amount of multimodal information it carries:
\begin{equation}\label{eq:lambda}
\lambda_e=\lambda_0\bigl[\widetilde{f}_{nc}(e)+\widetilde{f}_{bc}(e)+\delta\cdot M(e)\bigr],
\end{equation}
where $\widetilde{f}_{nc}(e)$ and $\widetilde{f}_{bc}(e)$ are the normalized degree centrality and betweenness centrality, respectively, and $M(e)$ is the modality richness. All three are normalized to $[0,1]$. The detailed formulas can be found in the appendix.

Constraining only the full entity representation is still not enough, because the multimodal projection layers themselves may also drift. We therefore further constrain the projected visual and textual representations of old entities:
\begin{equation}\label{eq:mod}
\mathcal{L}_{mod}=\sum_{e\in \mathcal{E}_{i-1}}\bigl\|\bar{\mathbf{v}}_e^{(i)}-\bar{\mathbf{v}}_e^{(i-1)}\bigr\|_2^2+\bigl\|\bar{\mathbf{w}}_e^{(i)}-\bar{\mathbf{w}}_e^{(i-1)}\bigr\|_2^2.
\end{equation}
At the same time, to prevent cross-modal relations from being disrupted during updating, we preserve the consistency of visual-text alignment:
\begin{equation}\label{eq:align}
\mathcal{L}_{align}=\sum_{e\in \mathcal{E}_{i-1}\cap \mathcal{E}_{vt}}\mathcal{D}\Bigl(\cos(\bar{\mathbf{v}}_e^{(i)},\bar{\mathbf{w}}_e^{(i)}),\ \cos(\bar{\mathbf{v}}_e^{(i-1)},\bar{\mathbf{w}}_e^{(i-1)})\Bigr),
\end{equation}
where $\mathcal{E}_{vt}$ is the set of entities that have both visual and textual modalities. For entities with only a single modality, the corresponding term in $\mathcal{L}_{align}$ is skipped, and in $\mathcal{L}_{mod}$ only the constraint for the available modality is computed.
This gives the entity-level preservation objective: $\mathcal{L}_{ent}=\mathcal{L}_{str}+\mathcal{L}_{mod}+\mathcal{L}_{align}$.

\subsubsection{Consistency of Relational Semantic Patterns}

We constrain relation representations from two perspectives: \emph{numerical stability} and \emph{stability of scoring patterns}. First, we directly enforce cross-step stability of the relation embeddings:
\begin{equation}\label{eq:remb}
\mathcal{L}_{r\text{-}emb}=\frac{1}{|\mathcal{R}_{i-1}|}\sum_{r\in \mathcal{R}_{i-1}}\mathcal{D}\bigl(\mathbf{r}^{(i)},\mathbf{r}^{(i-1)}\bigr).
\end{equation}

Second, we preserve the consistency of relational semantic patterns over the subset of replay triplets that involve old relations. The core idea is that the scores assigned by old relations to historical triplets should not change drastically after parameter updates. To this end, we define the replay subset of old relations as $\mathcal{T}_{rep}^{oldR}=\{(h,r,t)\in \mathcal{T}_{rep}\mid r\in \mathcal{R}_{i-1}\}$, and use the KGE scoring function to characterize pattern consistency:
\begin{equation}\label{eq:rpat}
\mathcal{L}_{r\text{-}pat}=\frac{1}{|\mathcal{T}_{rep}^{oldR}|}\sum_{(h,r,t)\in \mathcal{T}_{rep}^{oldR}}\mathcal{D}\Bigl(S^{(i)}(h,r,t),\ \mathrm{sg}\bigl(S^{(i-1)}(h,r,t)\bigr)\Bigr),
\end{equation}
where $S^{(i)}(h,r,t)$ denotes the TuckER score assigned by the model at snapshot $\mathcal{S}_i$ to $(h,r,t)$, and $\mathrm{sg}(\cdot)$ denotes the stop-gradient operation. Rather than simply forcing the numerical values of relation vectors to remain close, this term constrains the semantic patterns encoded by relations in the output space of the scoring function, so that the meaning captured by the \emph{head entity--relation--tail entity} combination remains consistent across snapshots. The relation-level preservation objective is therefore defined as
$\mathcal{L}_{rel}=\mathcal{L}_{r\text{-}emb}+\mathcal{L}_{r\text{-}pat}$.

\subsubsection{Modal Anchoring}

Freezing pretrained features at the input side alone is not enough to provide stable anchors, because once they pass through continuously updated parameters, the anchors themselves can still drift. We therefore construct anchors using the frozen model from the previous snapshot. For each old entity $e \in \mathcal{E}_{i-1}$, we set its structural embedding to zero, feed only the frozen pretrained modality tokens into the frozen encoder $\mathrm{Transformer}^{(i-1)}$, and take the output at the $[\mathrm{ENT}]$ position as the anchor $\mathbf{a}_e^{(i-1)}$. The detailed construction is given in the appendix.

We then use a projection head to constrain the current entity representation so that it does not move too far away from the anchor:
\begin{equation}\label{eq:anc}
\mathcal{L}_{anc}=\sum_{e\in \mathcal{E}_{i-1}}\mathcal{D}\bigl(\mathbf{P}(\mathbf{e}^{(i)}),\ \mathrm{sg}(\mathbf{Q}^{(i-1)}(\mathbf{a}_e^{(i-1)}))\bigr),
\end{equation}
where $\mathbf{P}(\cdot)$ and $\mathbf{Q}^{(i-1)}(\cdot)$ are the projection heads of the current model and the frozen model from the previous snapshot, respectively, both mapping from $\mathbb{R}^d$ to $\mathbb{R}^{d_p}$, and $\mathrm{sg}(\cdot)$ denotes the stop-gradient operation.
Combining the three parts above, the cross-modal knowledge preservation loss is defined as: $\mathcal{L}_{CMKP}=\mathcal{L}_{ent}+\mathcal{L}_{rel}+\mathcal{L}_{anc}$.

\subsection{Multimodal Contrastive Replay (MMCR) and Two-Stage Optimization}
\label{sec:mmcr}

\subsubsection{Multimodal Importance Sampling}

To preferentially retain old triplets that are structurally more important and richer in multimodal information, we define the importance score of any old triplet $(h,r,t)$ as
\begin{equation}\label{eq:importance}
w(h,r,t)=\frac{\deg(h)+\deg(t)}{2}\cdot \Bigl(1+\mathbb{I}[v_h]+\mathbb{I}[v_t]+\mathbb{I}[\mathrm{txt}_h]+\mathbb{I}[\mathrm{txt}_t]\Bigr),
\end{equation}
where $\deg(\cdot)$ is the degree of an entity in the current snapshot, and $\mathbb{I}[v_e]$ and $\mathbb{I}[\mathrm{txt}_e]$ indicate whether entity $e$ has visual and textual information, respectively.

Based on this score, we construct the replay buffer by probabilistic sampling, and allocate buffer capacity across different historical snapshots using a temporal proximity bias:
\begin{equation}\label{eq:alloc}
\mathrm{alloc}(j)=\frac{j+1}{\sum_{k=0}^{i-1}(k+1)}\cdot B,
\end{equation}
where $B$ is the total buffer size. The closer $j$ is to the current snapshot, the larger the replay capacity allocated to it.

\subsubsection{Contrastive Replay Objective}

When training on the current snapshot, we mix replay samples with newly added samples and feed them into the model together. To preserve semantic consistency between the old and new models on replay samples, we use a two-part loss.

First, for the set of unique entities $\mathcal{U}$ involved in replay triplets, we apply the InfoNCE loss to align their embeddings:
\begin{equation}\label{eq:rep_emb}
\mathcal{L}_{rep}^{emb}=-\frac{1}{|\mathcal{U}|}\sum_{e\in \mathcal{U}}\log \frac{\exp(\mathrm{sim}(\mathbf{e}^{(i)},\mathbf{e}^{(i-1)})/\tau)}{\sum_{e'\in \mathcal{U}}\exp(\mathrm{sim}(\mathbf{e}^{(i)},\mathbf{e'}^{(i-1)})/\tau)},
\end{equation}
where $\mathrm{sim}(\cdot,\cdot)$ is cosine similarity and $\tau$ is the temperature coefficient. Here $\mathcal{T}_{rep}$ is sampled only from historical snapshots $0,\ldots,i-1$, so every entity in $\mathcal{U}$ belongs to $\mathcal{E}_{i-1}$ and its representation $\mathbf{e}^{(i-1)}$ is always available from the frozen model of snapshot $i-1$. This loss enforces consistency between the old and new models in the representation space when encoding the same entity.

Second, we preserve the scoring consistency of old triplets under the new and old models:
\begin{equation}\label{eq:rep_score}
\mathcal{L}_{rep}^{score}=\frac{1}{|\mathcal{T}_{rep}|}\sum_{(h,r,t)\in \mathcal{T}_{rep}}\mathcal{D}\bigl(S^{(i)}(h,r,t),\ \mathrm{sg}(S^{(i-1)}(h,r,t))\bigr).
\end{equation}
Unlike $\mathcal{L}_{r\text{-}pat}$ in CMKP, $\mathcal{L}_{rep}^{score}$ is applied to \emph{all replay samples} rather than only the subset involving old relations. Moreover, the gradient from the current-model branch can update all model parameters, while the gradient of the old-model branch is blocked by $\mathrm{sg}$, so this term serves as a global constraint on score preservation. The multimodal contrastive replay loss is therefore defined as
\[
\mathcal{L}_{MMCR}=\mathcal{L}_{rep}^{emb}+\mathcal{L}_{rep}^{score}.
\]

\subsubsection{Two-Stage Optimization and Overall Objective}

To prevent insufficiently learned new entities from disrupting old knowledge during the early stage of training, we adopt a two-stage optimization strategy.

\textbf{Stage 1 (new knowledge adaptation):} We freeze the structural embeddings of old entities and the embeddings of old relations, and train only the newly introduced embeddings together with the shared parameters, including the Transformer encoder, the projection heads, and the parameters of KGE Models. At this stage, $\mathcal{L}_{mod}$ (Eq.~\ref{eq:mod}) and $\mathcal{L}_{align}$ (Eq.~\ref{eq:align}) are already activated to constrain modality drift caused by updates to the shared parameters, while any remaining drift is corrected in Stage 2.

\textbf{Stage 2 (global coordination):} We unfreeze all parameters and jointly activate the full $\mathcal{L}_{CMKP}$ and $\mathcal{L}_{MMCR}$ to coordinate old and new knowledge at the global level. The final training objective is written as
\begin{equation}\label{eq:total}
\mathcal{L}=\mathcal{L}_{kgr}+\lambda_{cmkp}\mathcal{L}_{CMKP}+\lambda_{rep}\mathcal{L}_{MMCR},
\end{equation}
where $\mathcal{L}_{kgr}$ is the link prediction loss, implemented as cross-entropy based on the TuckER score. At snapshot $\mathcal{S}_0$, both $\mathcal{L}_{CMKP}$ and $\mathcal{L}_{MMCR}$ are naturally zero, since there is no old knowledge to preserve. 

\begin{table*}[t]
    \centering
    \caption{Dataset statistics for all 9 continual MMKG benchmarks ($T\!=\!5$ snapshots). \#Ent and \#Rel denote cumulative entity/relation counts at each snapshot; \#Triples denotes the number of triples used at each snapshot after bridge augmentation, rather than only newly introduced triples (split into Train/Valid/Test at a 3:1:1 ratio).}
    \label{tab:dataset-stats}
    \setlength{\tabcolsep}{3.5pt}
    \footnotesize
    \begin{tabular}{@{}l l rrr rrr rrr rrr rrr@{}}
        \toprule
        \multirow{2}{*}{Dataset} & \multirow{2}{*}{Split}
        & \multicolumn{3}{c}{$\mathcal{S}_0$}
        & \multicolumn{3}{c}{$\mathcal{S}_1$}
        & \multicolumn{3}{c}{$\mathcal{S}_2$}
        & \multicolumn{3}{c}{$\mathcal{S}_3$}
        & \multicolumn{3}{c}{$\mathcal{S}_4$} \\
        \cmidrule(lr){3-5} \cmidrule(lr){6-8} \cmidrule(lr){9-11} \cmidrule(lr){12-14} \cmidrule(lr){15-17}
        & 
        & \#Ent & \#Rel & \#Triples
        & \#Ent & \#Rel & \#Triples
        & \#Ent & \#Rel & \#Triples
        & \#Ent & \#Rel & \#Triples
        & \#Ent & \#Rel & \#Triples \\
        \midrule

        \multirow{3}{*}{DB15K}
            & Entity
            & 4,494  & 216 & 50,149
            & 8,501  & 253 & 36,735
            & 10,504 & 261 & 12,255
            & 11,839 & 273 & 5,207
            & 12,842 & 279 & 2,012 \\
            & Higher
            & 1,884  & 136 & 8,616
            & 4,266  & 185 & 14,853
            & 6,813  & 206 & 19,806
            & 9,895  & 232 & 29,718
            & 12,842 & 279 & 28,290 \\
            & Equal
            & 4,332  & 181 & 19,808
            & 7,221  & 204 & 22,783
            & 9,459  & 229 & 22,783
            & 11,530 & 256 & 22,778
            & 12,842 & 279 & 11,451 \\
        \midrule

        \multirow{3}{*}{MKG-W}
            & Entity
            & 5,250  & 126 & 21,025
            & 9,930  & 150 & 15,891
            & 12,270 & 156 & 5,111
            & 13,830 & 166 & 2,651
            & 15,000 & 169 & 1,323 \\
            & Higher
            & 1,648  & 67 & 3,717
            & 3,891  & 67 & 6,411
            & 5,982  & 100 & 8,550
            & 9,166  & 129 & 12,825
            & 15,000 & 169 & 17,095 \\
            & Equal
            & 4,369  & 55 & 8,553
            & 6,094  & 98 & 9,847
            & 8,615  & 122 & 9,837
            & 10,345 & 135 & 9,834
            & 15,000 & 169 & 9,802 \\
        \midrule

        \multirow{3}{*}{MKG-Y}
            & Entity
            & 5,250  & 27 & 10,297
            & 9,930  & 27 & 10,560
            & 12,270 & 28 & 4,608
            & 13,830 & 28 & 2,386
            & 15,000 & 28 & 1,237 \\
            & Higher
            & 1,311  & 23 & 2,316
            & 3,355  & 28 & 3,998
            & 5,358  & 28 & 5,327
            & 8,739  & 28 & 7,991
            & 15,000 & 28 & 10,652 \\
            & Equal
            & 3,070  & 28 & 5,327
            & 5,489  & 28 & 6,131
            & 8,163  & 28 & 6,126
            & 10,961 & 28 & 6,126
            & 15,000 & 28 & 6,123 \\
        \bottomrule
    \end{tabular}
\end{table*}

\section{Experiment}
\subsection{Dataset}
We extend three public MMKG datasets, DB15K \cite{liu2019mmkg}, MKG-W\cite{xu2022relation}, and MKG-Y\cite{xu2022relation}, into nine datasets for continual learning; the statistics are reported in the table, and the construction details are provided in the appendix. For each dataset, we build an evolutionary sequence with $T=5$ steps under three partitioning strategies: progressive entity evolution, higher-increment evolution, and equal evolution. At each snapshot, the data are split into training, validation, and test sets at a ratio of 3:1:1, with 15\% bridging triples introduced to ensure connectivity across snapshots. Visual features are extracted using frozen BEiT\cite{bao2021beit}, textual features are extracted using frozen BERT\cite{devlin2018bert}, and the pretrained embeddings remain frozen.

\subsection{Baselines and Implementation Details}
Since no prior work directly addresses the continual multimodal KGR setting studied in this paper, we construct a comprehensive suite of more than 10 baselines by systematically adapting and combining methods from three related lines of research: unimodal KGE, continual KGE, and multimodal KGE. 
More specifically, we divide them into four groups based on two criteria: whether they are unimodal or multimodal, and whether they are non-continuous or continual.
\textbf{(i) Unimodal fine-tune} (lower bound): TransE~\cite{bordes2013translating}, DistMult~\cite{yang2014embedding}, ComplEx~\cite{trouillon2016complex}, RotatE~\cite {sun2019rotate}, and TuckER~\cite{balavzevic2019tucker}, each trained by naïve fine-tuning on successive snapshots without any forgetting mitigation;
\textbf{(ii) Unimodal + continual learning}: we equip the above models with two representative continual-learning strategies, EWC~\cite{kirkpatrick2017overcoming} and experience Replay, yielding eight additional variants (e.g., TransE+EWC, DistMult+Replay);
\textbf{(iii) Dedicated CKGE methods}: IncDE~\cite{liu2024towards} and FastKGE~\cite{liu2024fast}, which are originally designed for unimodal continual KGE and are used with their official implementations;
\textbf{(iv) Multimodal + continual learning}: to our knowledge, no existing method combines multimodal fusion with continual KGE. We therefore implement three multimodal base models: MMTransE, MMDistMult, and MMRotatE by extending their unimodal counterparts with a gated visual–textual fusion module, and further equip each with EWC or Replay, producing three fine-tune variants.
All baselines are evaluated under the same data splits and evaluation protocol for a fair comparison.

\noindent\textbf{Implementation details.}
MRCKG hyperparameters: $d=256$, batch size 1024, learning rate $5\times10^{-4}$, 200 epochs, and early stopping with patience 30. Experiments are conducted on a single NVIDIA V100-32GB GPU with PyTorch 2.4.0. For detailed hyper-parameter settings of each baseline, please refer to the supplementary material and model source code.

\subsection{Main Experimental Results}
Table~\ref{tab:main-results} presents a comparison of the Avg MRR and Hits@10 for all 18 methods across 9 benchmarks (where each metric represents the mean value obtained by the final model on test\_0 through test\_4). More detailed and complete results of the main experiments can be found in the Appendix.

As shown in Table~\ref{tab:main-results}, MRCKG achieves the best MRR on all nine benchmarks. On DB15K-Entity, for example, MRCKG improves MRR by 13.4\% over the strongest continual baseline, IncDE. In contrast, simple gated fusion is almost ineffective under continual learning: the gap between MMTransE+FT and TransE+FT is no more than 0.01, indicating that multimodal gains cannot be realized without a dedicated continual-learning mechanism. Meanwhile, the effectiveness of the generic continual-learning strategy EWC depends heavily on the underlying model. It is consistently beneficial for ComplEx, but causes a 51.4\% drop on TuckER, suggesting poor transferability to the KGR setting. Across datasets, MRCKG remains superior on the dense graph DB15K, the medium-density graph MKG-W, and the sparse graph MKG-Y. The largest gain appears on MKG-W-Higher, reaching 30.6\%, which shows that the multimodal anchor mechanism provides robust benefits across different graph structures.

\begin{table*}[htbp]
    \centering
    \caption{Average MRR and Hits@10 comparison on all nine benchmarks. The best result in each column is shown in \textbf{bold}, and the second best is \underline{underlined}.}
    \label{tab:main-results}
    \setlength{\tabcolsep}{2.2pt}
    \footnotesize
    \resizebox{0.95\textwidth}{!}{%
    \begin{tabular}{@{}ll*{18}{c}@{}}
        \toprule
                &                            & \multicolumn{6}{c}{DB15K} & \multicolumn{6}{c}{MKG-W} & \multicolumn{6}{c}{MKG-Y} \\
        \cmidrule(lr){3-8} \cmidrule(lr){9-14} \cmidrule(lr){15-20}
                &                            & \multicolumn{2}{c}{Entity} & \multicolumn{2}{c}{Higher} & \multicolumn{2}{c}{Equal}
                                             & \multicolumn{2}{c}{Entity} & \multicolumn{2}{c}{Higher} & \multicolumn{2}{c}{Equal}
                                             & \multicolumn{2}{c}{Entity} & \multicolumn{2}{c}{Higher} & \multicolumn{2}{c}{Equal} \\
        \cmidrule(lr){3-4} \cmidrule(lr){5-6} \cmidrule(lr){7-8}
        \cmidrule(lr){9-10} \cmidrule(lr){11-12} \cmidrule(lr){13-14}
        \cmidrule(lr){15-16} \cmidrule(lr){17-18} \cmidrule(lr){19-20}
        Setting & Method                     & MRR     & H@10   & MRR     & H@10   & MRR     & H@10
                                             & MRR     & H@10   & MRR     & H@10   & MRR     & H@10
                                             & MRR     & H@10   & MRR     & H@10   & MRR     & H@10 \\
        \midrule
        \multirow{4}{*}{\shortstack[l]{Unimodal\\+FT}}
                & TransE+FT                  & 0.1085  & 0.2355 & 0.0987  & 0.2271 & 0.0913  & 0.2160 & 0.0956  & 0.1909 & 0.1024  & 0.2289 & 0.0867  & 0.1919 & 0.1243  & 0.2212 & 0.1587  & 0.2962 & 0.1462  & 0.2697 \\
                & DistMult+FT                & 0.1186  & 0.2092 & 0.1063  & 0.1988 & 0.0978  & 0.1880 & 0.1043  & 0.1693 & 0.1098  & 0.1995 & 0.0942  & 0.1695 & 0.1356  & 0.1961 & 0.1725  & 0.2617 & 0.1589  & 0.2382 \\
                & ComplEx+FT                 & 0.1644  & 0.2348 & 0.1512  & 0.2289 & 0.1387  & 0.2159 & 0.1425  & 0.1872 & 0.1536  & 0.2260 & 0.1308  & 0.1905 & 0.1872  & 0.2192 & 0.2316  & 0.2845 & 0.2153  & 0.2614 \\
                & TuckER+FT                  & 0.1618  & 0.2151 & 0.1489  & 0.2098 & 0.1354  & 0.1962 & 0.1397  & 0.1709 & 0.1503  & 0.2058 & 0.1276  & 0.1730 & 0.1835  & 0.2000 & 0.2274  & 0.2600 & 0.2108  & 0.2382 \\
        \midrule
        \multirow{6}{*}{\shortstack[l]{Unimodal\\+CL}}
                & TransE+EWC                 & 0.1088  & 0.2276 & 0.0994  & 0.2204 & 0.0921  & 0.2100 & 0.0968  & 0.1863 & 0.1037  & 0.2234 & 0.0879  & 0.1876 & 0.1258  & 0.2158 & 0.1604  & 0.2886 & 0.1478  & 0.2628 \\
                & TransE+Replay              & 0.1155  & 0.2386 & 0.1067  & 0.2336 & 0.0982  & 0.2211 & 0.1027  & 0.1952 & 0.1095  & 0.2330 & 0.0938  & 0.1976 & 0.1315  & 0.2228 & 0.1672  & 0.2970 & 0.1543  & 0.2709 \\
                & DistMult+EWC               & 0.1224  & 0.2074 & 0.1102  & 0.1979 & 0.1015  & 0.1875 & 0.1074  & 0.1674 & 0.1127  & 0.1967 & 0.0973  & 0.1682 & 0.1389  & 0.1930 & 0.1758  & 0.2562 & 0.1624  & 0.2339 \\
                & ComplEx+EWC                & 0.1679  & 0.2361 & 0.1538  & 0.2292 & 0.1419  & 0.2175 & 0.1486  & 0.1922 & 0.1572  & 0.2277 & 0.1342  & 0.1925 & 0.1907  & 0.2199 & 0.2362  & 0.2856 & 0.2197  & 0.2626 \\
                & RotatE+EWC                 & 0.1225  & 0.1755 & 0.1109  & 0.1684 & 0.1028  & 0.1605 & 0.1098  & 0.1447 & 0.1156  & 0.1706 & 0.0986  & 0.1441 & 0.1412  & 0.1659 & 0.1793  & 0.2209 & 0.1659  & 0.2020 \\
                & TuckER+EWC                 & 0.0786  & 0.1243 & 0.0723  & 0.1212 & 0.0654  & 0.1127 & 0.0679  & 0.0988 & 0.0712  & 0.1160 & 0.0598  & 0.0965 & 0.0914  & 0.1185 & 0.1178  & 0.1602 & 0.1087  & 0.1461 \\
        \midrule
        \multirow{2}{*}{\shortstack[l]{Dedicated\\CKGE}}
                & IncDE                      & \underline{0.2518} & \underline{0.3617} & \underline{0.2347} & \underline{0.3567} & \underline{0.2215} & \underline{0.3472} & 0.2098 & 0.2768 & \underline{0.2261} & \underline{0.3344} & \underline{0.2034} & \underline{0.2987} & \underline{0.2394} & \underline{0.2827} & \underline{0.3186} & \underline{0.3931} & \underline{0.3057} & \underline{0.3707} \\
                & FastKGE                    & 0.2387  & 0.3428 & 0.2058  & 0.3128 & 0.2082  & 0.3263 & \underline{0.2137} & \underline{0.2819} & 0.2173  & 0.3214 & 0.1697  & 0.2492 & 0.2347  & 0.2771 & 0.3082  & 0.3803 & 0.2842  & 0.3446 \\
        \midrule
        \multirow{3}{*}{\shortstack[l]{Multimodal\\+FT}}
                & MMTransE+FT                & 0.1084  & 0.2352 & 0.0976  & 0.2245 & 0.0905  & 0.2140 & 0.0963  & 0.1922 & 0.1038  & 0.2320 & 0.0858  & 0.1899 & 0.1237  & 0.2201 & 0.1579  & 0.2946 & 0.1453  & 0.2680 \\
                & MMDistMult+FT              & 0.1326  & 0.1970 & 0.1198  & 0.1887 & 0.1104  & 0.1788 & 0.1175  & 0.1606 & 0.1247  & 0.1908 & 0.1064  & 0.1612 & 0.1498  & 0.1825 & 0.1896  & 0.2422 & 0.1752  & 0.2212 \\
                & MMRotatE+FT                & 0.1117  & 0.1603 & 0.1024  & 0.1558 & 0.0937  & 0.1466 & 0.1008  & 0.1331 & 0.1074  & 0.1588 & 0.0893  & 0.1307 & 0.1279  & 0.1505 & 0.1635  & 0.2018 & 0.1508  & 0.1840 \\
        \midrule
        \multirow{2}{*}{\shortstack[l]{Multimodal\\+CL}}
                & MMTransE+EWC               & 0.1136  & 0.2385 & 0.1042  & 0.2319 & 0.0958  & 0.2192 & 0.1014  & 0.1959 & 0.1089  & 0.2355 & 0.0912  & 0.1953 & 0.1302  & 0.2241 & 0.1657  & 0.2992 & 0.1529  & 0.2729 \\
                & MMTransE+Replay            & 0.1170  & 0.2589 & 0.1083  & 0.2540 & 0.0991  & 0.2390 & 0.1052  & 0.2142 & 0.1125  & 0.2564 & 0.0946  & 0.2135 & 0.1341  & 0.2433 & 0.1704  & 0.3243 & 0.1572  & 0.2957 \\
        \midrule
        \multicolumn{2}{@{}l}{\textbf{MRCKG (Ours)}}
                & \textbf{0.2856} & \textbf{0.4102} & \textbf{0.2614} & \textbf{0.3973} & \textbf{0.2305} & \textbf{0.3613}
                & \textbf{0.2712} & \textbf{0.3578} & \textbf{0.2953} & \textbf{0.4367} & \textbf{0.2183} & \textbf{0.3206}
                & \textbf{0.2518} & \textbf{0.2973} & \textbf{0.3285} & \textbf{0.4053} & \textbf{0.3146} & \textbf{0.3815} \\
        \bottomrule
    \end{tabular}%
    }
\end{table*}

\subsection{Ablation Studies}
We further extend the ablation study from component-level analysis to modality-level analysis on DB15K-Entity. Table~\ref{tab:ablation} reports both the effects of removing each core component and the results of discarding either the visual or textual modality. Four observations can be drawn: (1) Removing CMKP causes the largest performance drop, reducing MRR by 10.8\% and severely worsening BWT, indicating that CMKP is the key to resisting forgetting. (2) Removing MMCR also leads to a clear degradation, confirming the role of contrastive replay in balancing old and new knowledge. (3) Comparing w/o MSCL with w/o Prog., the latter performs worse and yields lower BWT, showing that curriculum scoring and progressive training play complementary roles within MSCL: the former prioritizes samples, while the latter regulates the training pace. (4) In the modality ablation, removing the visual modality leads to a larger drop in MRR and Hits@1, whereas removing the textual modality causes a relatively larger decline in Hits@10 and a slightly worse BWT. This suggests that visual signals contribute more to precise top-rank prediction, while textual semantics provide complementary support for broader candidate coverage and continual retention.
\begin{table}[htbp]
    \centering\small
    \caption{Ablation study on DB15K-Entity, including component ablation and modality ablation.}
    \label{tab:ablation}
    \begin{tabular}{@{}lcccc@{}}
        \toprule
        Variant      & MRR             & H@1             & H@10            & BWT               \\
        \midrule
        Full         & \textbf{0.2856} & \textbf{0.2213} & \textbf{0.4102} & \textbf{$-$0.068} \\
        \midrule
        w/o CMKP     & 0.2549          & 0.1988          & 0.3673          & $-$0.126          \\
        w/o MMCR     & 0.2683          & 0.2056          & 0.3875          & $-$0.095          \\
        w/o MSCL     & 0.2715          & 0.2075          & 0.3914          & $-$0.085          \\
        w/o Prog.    & 0.2639          & 0.2021          & 0.3749          & $-$0.102          \\
        w/o Visual   & 0.2671          & 0.2003          & 0.3931          & $-$0.073          \\
        w/o Textual  & 0.2807          & 0.2111          & 0.3842          & $-$0.074          \\
        \bottomrule
    \end{tabular}
\end{table}
\begin{figure}[htbp]
    \centering
    \includegraphics[width=0.8\columnwidth]{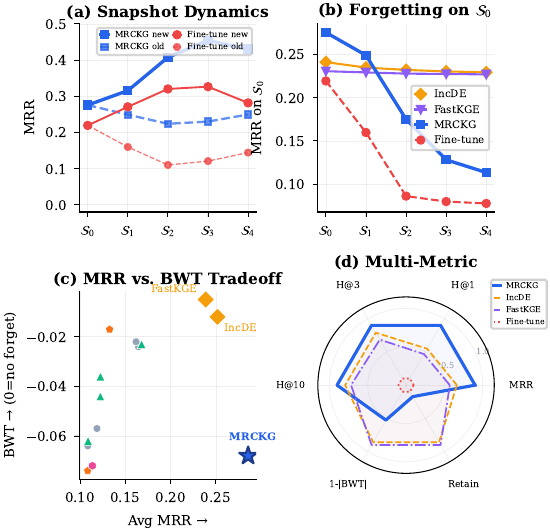}
    \caption{Analysis on DB15K-Entity. (a) Per-snapshot MRR on new vs.\ old triples; (b) $\mathcal{S}_0$ forgetting curves; (c) MRR--BWT Pareto front; (d) multi-metric radar plot.}
    \Description{Four plots compare continual-learning behavior on DB15K-Entity: MRR for new and old triples, forgetting on the first snapshot, the MRR versus backward-transfer frontier, and a radar chart of normalized performance metrics.}
    \label{fig:analysis}
\end{figure}
\subsection{The Forgetting Curve and Snapshot Dynamic Analysis}
Figure~\ref{fig:analysis} takes DB15K-Entity as an example to examine the continual learning behavior of MRCKG from four perspectives. Figure (a) shows that MRCKG reaches a much higher peak $\mathrm{MRR}_{\mathrm{new}}$ than Fine-tune, while its drop in $\mathrm{MRR}_{\mathrm{old}}$ is far smaller, suggesting a better balance between plasticity and stability. In the forgetting curve on $\mathcal{S}_0$ in Figure~\ref{fig:analysis}(b), IncDE and FastKGE achieve the highest retention thanks to their dedicated anti-forgetting designs. MRCKG still faces forgetting on a single test set, but its overall $\mathrm{MRR}_{\mathrm{old}}$ remains clearly better than that of Fine-tune.

Figures~\ref{fig:analysis}(c) and (d) further show that MRCKG lies on the high-MRR frontier and outperforms IncDE on multiple metrics. MRCKG is weaker than IncDE and FastKGE on BWT and Retain because multimodal joint optimization involves a larger parameter space: it improves accuracy, but also increases the risk of overwriting old knowledge. By contrast, IncDE and FastKGE encode only structural information, which naturally favors stability. This trade-off between accuracy and forgetting is common in multimodal continual learning and points to the need for more precise memory preservation mechanisms in future work.

\begin{figure}[htbp]
    \centering
    \includegraphics[width=0.8\columnwidth]{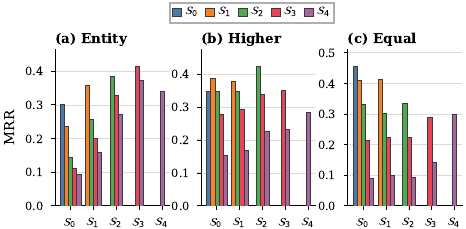}
    \caption{Per-snapshot MRR bar plots of MRCKG on the three DB15K settings.}
    \Description{Three grouped bar charts show MRR for snapshots zero through four after successive training stages under the Entity, Higher, and Equal DB15K splits.}
    \label{fig:temporal-higher}
\end{figure}
\subsection{Performance of MRCKG in Each Snapshot}
Figure~\ref{fig:temporal-higher} shows the per snapshot evaluation results of MRCKG on the three DB15K splits. In each group of bars, the models are ordered from left to right by the snapshot they have been trained up to. The Entity setting (Figure ~\ref{fig:temporal-higher}(a)) shows a clear pattern of learning new knowledge while forgetting old knowledge. As training goes on, the MRR on earlier snapshots keeps dropping, while performance on later snapshots keeps improving. In the Higher setting (Figure~\ref{fig:temporal-higher}(b)), $\mathcal{S}_0$ shows positive backward transfer. Early training even improves performance on old knowledge, which suggests that old and new knowledge can help each other. But as training continues, forgetting still cannot be avoided. The Equal setting (Figure~\ref{fig:temporal-higher}(c)) shows the strongest forgetting. With an even split, competition between snapshots becomes more intense.
\subsection{Error Type Analysis}
\begin{figure}[htbp]
    \centering
    \includegraphics[width=0.8\columnwidth]{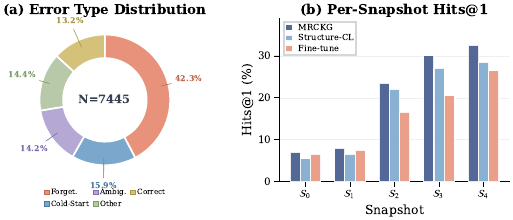}
    \caption{(a) Error type distribution of MRCKG; (b) Hits@1 of three methods across snapshots.}
    \Description{A donut chart partitions MRCKG predictions into forgetting, cold-start, cross-modal ambiguity, other, and correct outcomes. A grouped bar chart compares MRCKG, Structure-CL, and fine-tuning Hits at one across five snapshots.}
    \label{fig:case-study}
\end{figure}

To better understand the model’s behavior, we categorized the errors in all 7,445 test predictions made by MRCKG on DB15K-Entity, as shown in Figure~\ref{fig:case-study}(a). The distribution reveals three main findings. First, forgetting errors account for the largest share, indicating that degradation of old knowledge remains the main bottleneck in continual learning, even with multimodal anti-forgetting mechanisms. This also highlights a clear direction for future improvement. Second, cross-modal ambiguity errors show that similar multimodal features can cause entity confusion, suggesting the need for finer-grained cross-modal discrimination. Third, cold-start errors are mainly associated with newly introduced entities in later snapshots, reflecting the inherent limitation of insufficient training data.

Figure~\ref{fig:case-study}(b) compares Hits@1 across snapshots for three methods. \textbf{Structure-CL} is an ablation variant of MRCKG: it keeps all continual learning components (MSCL, CMKP, and MMCR) but removes all multimodal inputs by zeroing out visual and textual tokens, reducing the model to a purely structural continual learning method. \textbf{Fine-tune}, in contrast, uses full multimodal information but no continual learning mechanism, and is directly fine-tuned at each step. MRCKG achieves the best performance on every snapshot. The gap among the three methods is small in the early snapshots, suggesting that old knowledge degradation is a common challenge in continual learning. As training proceeds, however, MRCKG’s advantage becomes increasingly clear, with a particularly large lead in the later snapshots. This shows that multimodal information provides effective semantic support for learning new knowledge.

\section{Conclusion}
This paper presents MRCKG, a unified framework for the CMMKGR task, with three core modules: MSCL, CMKP, and MMCR. MSCL builds a curriculum score from structural connectivity, multimodal compatibility, and modality richness, and organizes training samples from easy to hard. CMKP defines a unified memory-preservation objective based on entity embedding stability, consistency between relation embeddings and scoring patterns, and modality anchoring. MMCR further strengthens past knowledge through multimodal-aware contrastive replay. Across systematic comparisons on multiple benchmarks and baselines, MRCKG consistently achieves the best MRR and clearly outperforms the strongest task-specific CKGE baseline. The results also show that general continual learning methods depend heavily on the backbone model and bring only limited gains, while simple multimodal fusion offers almost no benefit in continual learning settings. By contrast, the three modules in MRCKG work together to make multimodal information serve as a real semantic anchor. We believe the inherent stability of pretrained multimodal features provides a reliable reference for updating structural embeddings in continual learning, opening up a new direction for the continual evolution of MMKGs. To support the advancement of CMMKGR, our code is available at: https://anonymous.4open.science/r/MRCKG-AC21.

\begin{acks}
This work was supported by the Key Program of the National Natural Science Foundation of China (Grant No. 62436006) and the Key Research and Development Program of the Tibet Autonomous Region (Grant No. XZ202601ZY0087).
\end{acks}

\bibliographystyle{ACM-Reference-Format}
\bibliography{sample-base}

\clearpage
\appendix
\input{appendix}

\end{document}

%% file: appendix.tex

\section{Appendices}
\subsection{Additional Methodological Details}

\subsubsection{Complete Definition of Multimodal Compatibility $c_{mm}$}\label{app:cmm}

Let $\mathcal{U}_{new}(h,t)=\{u\in\{h,t\}\mid u\notin \mathcal{E}_{old}\}$ denote the set of newly introduced endpoints in the triplet $(h,t)$. The complete piecewise definition of multimodal compatibility is given by
\begin{equation}
    c_{mm}(h,t\mid \mathcal{E}_{old})=
    \begin{cases}
        \displaystyle\max_{u\in\mathcal{U}_{new}}\max_{e'\in \mathcal{E}_{old}}\mathrm{sim}_{mm}(u,e'), & \mathcal{U}_{new}\neq\varnothing, \\
        0,                                                                                              & \mathcal{U}_{new}=\varnothing,
    \end{cases}
\end{equation}
where $\mathrm{sim}_{mm}(u,e')=\eta_v\cdot \cos(\mathbf{v}_u^{pt},\mathbf{v}_{e'}^{pt})+\eta_t\cdot \cos(\mathbf{w}_u^{pt},\mathbf{w}_{e'}^{pt})$. Here, $\mathbf{v}_e^{pt}$ and $\mathbf{w}_e^{pt}$ are the mean-pooled outputs of the BEiT codebook vectors and BERT word embeddings, respectively, obtained directly from the frozen pretrained encoders rather than from any learnable projection layer. Since these pretrained features are available before training and do not depend on model parameters, they allow semantic similarity to be estimated reliably even when newly introduced entities have not yet been trained.

\textbf{Handling missing modalities.} If either $u$ or $e'$ lacks a given modality, the corresponding cosine similarity term is set to 0, and $\eta_v,\eta_t$ are renormalized over the available modalities. For example, if both entities have text features but $u$ lacks visual features, only the textual term is retained and $\eta_t$ is set to 1.

\textbf{Why compatibility is computed only for new endpoints.} To avoid artificially inflated $c_{mm}$ values caused by self-matching of old endpoints (for which cosine similarity is always 1), this term is computed only for newly introduced endpoints. When both ends of a triplet are old entities ($\mathcal{U}_{new}=\varnothing$), structural connectivity is already captured by $c_{str}$, so no additional multimodal compatibility term is needed.

\textbf{Computational efficiency.} In practice, an approximate nearest neighbor index (e.g., FAISS) is built over $\mathcal{E}_{old}$, reducing the cost of pairwise similarity search from $O(|\mathcal{E}_{old}|)$ to approximately $O(\log|\mathcal{E}_{old}|)$.

\subsubsection{Centrality Measures}\label{app:centrality}

The two centrality measures used in the entity importance weight $\lambda_e$ are defined as follows.

\textbf{Degree Centrality.} $f_{nc}(e)=\deg(e)$, i.e., the degree of entity $e$ in the current snapshot graph, defined as the number of edges directly connected to it. Degree centrality reflects the local connectivity importance of an entity: entities with higher degree participate in more triplets in the graph, so changes in their embeddings have a larger impact on the overall graph representation. As a result, they require stronger stability constraints in continual learning.

\textbf{Betweenness Centrality.} $f_{bc}(e)=\sum_{s\neq e \neq t}\frac{\sigma_{st}(e)}{\sigma_{st}}$, where $\sigma_{st}$ is the total number of shortest paths between entities $s$ and $t$, and $\sigma_{st}(e)$ is the number of those paths that pass through $e$. Betweenness centrality measures the bridging role of an entity in information propagation across the graph: entities with high betweenness often connect different substructures, and shifts in their embeddings may simultaneously degrade representations in multiple local regions.

Both centrality measures are normalized to $[0,1]$ via min--max normalization before use:
\[
    \widetilde{f}_{nc}(e)=\frac{f_{nc}(e)-f_{nc}^{\min}}{f_{nc}^{\max}-f_{nc}^{\min}},\qquad
    \widetilde{f}_{bc}(e)=\frac{f_{bc}(e)-f_{bc}^{\min}}{f_{bc}^{\max}-f_{bc}^{\min}}.
\]
This normalization places degree centrality, betweenness centrality, and modality richness $M(e)$ on the same numerical scale, preventing the typically larger values of betweenness centrality from dominating the weight assignment.

\subsubsection{Detailed Construction of Modality Anchors}\label{app:anchor}

For each old entity $e \in \mathcal{E}_{i-1}$, the modality anchor is constructed as follows. The structural embedding is set to zero ($\mathbf{s}_e = \mathbf{0}$), and only the frozen pretrained modality tokens are fed into the encoder with parameters $\theta^{(i-1)}$ frozen from the previous snapshot:
\begin{equation}
    \mathbf{a}_e^{(i-1)} = \mathrm{Transformer}^{(i-1)}\bigl([\mathrm{ENT}] \oplus \mathbf{0} \oplus \hat{\mathbf{v}}_{e,1:m}^{pt} \oplus \hat{\mathbf{w}}_{e,1:n}^{pt}\bigr)\Big|_{[\mathrm{ENT}]},
\end{equation}
where $\hat{\mathbf{v}}_{e,1:m}^{pt}$ and $\hat{\mathbf{w}}_{e,1:n}^{pt}$ are the frozen pretrained tokens mapped through the projection layer from the previous step, and $\mathrm{Transformer}^{(i-1)}$ denotes the encoder frozen at the end of snapshot $i-1$. Since $\theta^{(i-1)}$ is no longer updated at the current step, $\mathbf{a}_e^{(i-1)}$ remains fixed throughout training at snapshot $i$, thereby serving as a true cross-temporal anchor.

\textbf{Motivation for using projection layers.} In the anchor loss $\mathcal{L}_{anc}$, the projection layers $\mathbf{P}(\cdot)$ and $\mathbf{Q}^{(i-1)}(\cdot)$ are used instead of imposing the constraint directly in the original representation space. The reason is that the full entity representation $\mathbf{e}^{(i)}$ needs to encode both structural topology and multimodal semantics. Directly pulling it toward a purely modality-based anchor would overly restrict the plasticity of the structural embedding. The projection layer $\mathbf{P}$ therefore allows the model to learn a semantic subspace in which alignment with the anchor is preserved, while still permitting the structural information in the original space to be updated freely.



\subsection{Full Experimental Results}
\label{sec:appendix-results}

Tables~\ref{tab:appendix-db15k},~\ref{tab:appendix-mkgw}, and~\ref{tab:appendix-mkgy} present the complete results of all methods on the DB15K, MKG-W, and MKG-Y benchmark families under the Entity, Higher, and Equal splits. Five metrics are reported: Avg MRR, Avg Hits@1, Avg Hits@3, Avg Hits@10, and BWT. All values are computed by the final model (after training on $\mathcal{S}_4$) and averaged over the test sets of snapshots $\mathcal{S}_0$ through $\mathcal{S}_4$.

\begin{table*}[h]
    \centering
    \caption{Full results on DB15K benchmarks (Avg MRR / H@1 / H@3 / H@10 / BWT). Best \textbf{bold}, second best \underline{underlined}.}
    \label{tab:appendix-db15k}
    \setlength{\tabcolsep}{1.8pt}
    \scriptsize
    \begin{tabular}{@{}l|ccccc|ccccc|ccccc@{}}
        \toprule
        Method          & \multicolumn{5}{c}{Entity} & \multicolumn{5}{c}{Higher} & \multicolumn{5}{c}{Equal} \\
        \cmidrule(lr){2-6} \cmidrule(lr){7-11} \cmidrule(lr){12-16}
                        & MRR & H@1 & H@3 & H@10 & BWT & MRR & H@1 & H@3 & H@10 & BWT & MRR & H@1 & H@3 & H@10 & BWT \\
        \midrule
        TransE+FT & 0.1085 & 0.0427 & 0.1331 & 0.2355 & $-$0.064 & 0.0987 & 0.0350 & 0.1247 & 0.2271 & $-$0.067 & 0.0913 & 0.0334 & 0.1165 & 0.2160 & $-$0.074 \\
        DistMult+FT & 0.1186 & 0.0751 & 0.1274 & 0.2092 & $-$0.057 & 0.1063 & 0.0606 & 0.1176 & 0.1988 & $-$0.060 & 0.0978 & 0.0576 & 0.1093 & 0.1880 & $-$0.066 \\
        ComplEx+FT & 0.1644 & 0.1275 & 0.1798 & 0.2348 & $-$0.024 & 0.1512 & 0.1055 & 0.1703 & 0.2289 & $-$0.025 & 0.1387 & 0.1000 & 0.1578 & 0.2159 & $-$0.028 \\
        TuckER+FT & 0.1618 & \underline{0.1338} & 0.1724 & 0.2151 & $-$0.022 & 0.1489 & \underline{0.1108} & 0.1634 & 0.2098 & $-$0.023 & 0.1354 & \underline{0.1041} & 0.1500 & 0.1962 & $-$0.025 \\
        \midrule
        TransE+EWC & 0.1088 & 0.0477 & 0.1301 & 0.2276 & $-$0.062 & 0.0994 & 0.0392 & 0.1224 & 0.2204 & $-$0.065 & 0.0921 & 0.0376 & 0.1145 & 0.2100 & $-$0.071 \\
        TransE+Replay & 0.1155 & 0.0480 & 0.1435 & 0.2386 & $-$0.058 & 0.1067 & 0.0399 & 0.1365 & 0.2336 & $-$0.061 & 0.0982 & 0.0380 & 0.1269 & 0.2211 & $-$0.067 \\
        DistMult+EWC & 0.1224 & 0.0800 & 0.1359 & 0.2074 & $-$0.044 & 0.1102 & 0.0648 & 0.1260 & 0.1979 & $-$0.046 & 0.1015 & 0.0617 & 0.1172 & 0.1875 & $-$0.051 \\
        ComplEx+EWC & \underline{0.1679} & 0.1313 & \underline{0.1854} & 0.2361 & $-$0.023 & \underline{0.1538} & 0.1082 & \underline{0.1749} & 0.2292 & $-$0.024 & \underline{0.1419} & 0.1032 & \underline{0.1630} & 0.2175 & $-$0.026 \\
        RotatE+EWC & 0.1225 & 0.0952 & 0.1310 & 0.1755 & $-$0.036 & 0.1109 & 0.0776 & 0.1222 & 0.1684 & $-$0.038 & 0.1028 & 0.0743 & 0.1143 & 0.1605 & $-$0.041 \\
        TuckER+EWC & 0.0786 & 0.0549 & 0.0873 & 0.1243 & \textbf{$-$0.015} & 0.0723 & 0.0454 & 0.0827 & 0.1212 & \textbf{$-$0.016} & 0.0654 & 0.0425 & 0.0755 & 0.1127 & \textbf{$-$0.017} \\
        \midrule
        MMTransE+FT & 0.1084 & 0.0372 & 0.1433 & 0.2352 & $-$0.074 & 0.0976 & 0.0301 & 0.1329 & 0.2245 & $-$0.078 & 0.0905 & 0.0289 & 0.1244 & 0.2140 & $-$0.085 \\
        MMDistMult+FT & 0.1326 & 0.0984 & 0.1429 & 0.1970 & \underline{$-$0.017} & 0.1198 & 0.0800 & 0.1330 & 0.1887 & \underline{$-$0.018} & 0.1104 & 0.0762 & 0.1237 & 0.1788 & \underline{$-$0.020} \\
        MMRotatE+FT & 0.1117 & 0.0861 & 0.1178 & 0.1603 & $-$0.040 & 0.1024 & 0.0710 & 0.1112 & 0.1558 & $-$0.042 & 0.0937 & 0.0672 & 0.1028 & 0.1466 & $-$0.046 \\
        \midrule
        MMTransE+EWC & 0.1136 & 0.0409 & 0.1520 & 0.2385 & $-$0.072 & 0.1042 & 0.0338 & 0.1436 & 0.2319 & $-$0.076 & 0.0958 & 0.0321 & 0.1333 & 0.2192 & $-$0.083 \\
        MMTransE+Replay & 0.1170 & 0.0415 & 0.1519 & \underline{0.2589} & $-$0.064 & 0.1083 & 0.0346 & 0.1448 & \underline{0.2540} & $-$0.067 & 0.0991 & 0.0327 & 0.1338 & \underline{0.2390} & $-$0.074 \\
        \midrule
        \textbf{MRCKG} & \textbf{0.2856} & \textbf{0.2213} & \textbf{0.3148} & \textbf{0.4102} & $-$0.068 & \textbf{0.2614} & \textbf{0.1879} & \textbf{0.2968} & \textbf{0.3973} & $-$0.071 & \textbf{0.2305} & \textbf{0.1587} & \textbf{0.2642} & \textbf{0.3613} & $-$0.078 \\
        \bottomrule
    \end{tabular}
\end{table*}
\begin{table*}[h]
    \centering
    \caption{Full results on MKG-W benchmarks (Avg MRR / H@1 / H@3 / H@10 / BWT). Best \textbf{bold}, second best \underline{underlined}.}
    \label{tab:appendix-mkgw}
    \setlength{\tabcolsep}{1.8pt}
    \scriptsize
    \begin{tabular}{@{}l|ccccc|ccccc|ccccc@{}}
        \toprule
        Method          & \multicolumn{5}{c}{Entity} & \multicolumn{5}{c}{Higher} & \multicolumn{5}{c}{Equal} \\
        \cmidrule(lr){2-6} \cmidrule(lr){7-11} \cmidrule(lr){12-16}
                        & MRR & H@1 & H@3 & H@10 & BWT & MRR & H@1 & H@3 & H@10 & BWT & MRR & H@1 & H@3 & H@10 & BWT \\
        \midrule
        TransE+FT & 0.0956 & 0.0376 & 0.1138 & 0.1909 & $-$0.061 & 0.1024 & 0.0351 & 0.1269 & 0.2289 & $-$0.066 & 0.0867 & 0.0314 & 0.1074 & 0.1919 & $-$0.072 \\
        DistMult+FT & 0.1043 & 0.0660 & 0.1087 & 0.1693 & $-$0.054 & 0.1098 & 0.0605 & 0.1191 & 0.1995 & $-$0.059 & 0.0942 & 0.0549 & 0.1022 & 0.1695 & $-$0.064 \\
        ComplEx+FT & 0.1425 & 0.1105 & 0.1512 & 0.1872 & $-$0.023 & 0.1536 & 0.1036 & 0.1697 & 0.2260 & $-$0.025 & 0.1308 & 0.0933 & 0.1445 & 0.1905 & $-$0.027 \\
        TuckER+FT & 0.1397 & 0.1155 & 0.1444 & 0.1709 & $-$0.021 & 0.1503 & \underline{0.1081} & 0.1617 & 0.2058 & $-$0.023 & 0.1276 & \underline{0.0971} & 0.1373 & 0.1730 & $-$0.025 \\
        \midrule
        TransE+EWC & 0.0968 & 0.0424 & 0.1123 & 0.1863 & $-$0.059 & 0.1037 & 0.0396 & 0.1252 & 0.2234 & $-$0.064 & 0.0879 & 0.0355 & 0.1062 & 0.1876 & $-$0.069 \\
        TransE+Replay & 0.1027 & 0.0427 & 0.1238 & 0.1952 & $-$0.055 & 0.1095 & 0.0396 & 0.1374 & 0.2330 & $-$0.060 & 0.0938 & 0.0359 & 0.1177 & 0.1976 & $-$0.065 \\
        DistMult+EWC & 0.1074 & 0.0702 & 0.1157 & 0.1674 & $-$0.042 & 0.1127 & 0.0641 & 0.1264 & 0.1967 & $-$0.045 & 0.0973 & 0.0585 & 0.1091 & 0.1682 & $-$0.049 \\
        ComplEx+EWC & \underline{0.1486} & \underline{0.1162} & \underline{0.1592} & 0.1922 & $-$0.022 & \underline{0.1572} & 0.1070 & \underline{0.1753} & 0.2277 & $-$0.024 & \underline{0.1342} & 0.0966 & \underline{0.1497} & 0.1925 & $-$0.026 \\
        RotatE+EWC & 0.1098 & 0.0853 & 0.1139 & 0.1447 & $-$0.034 & 0.1156 & 0.0782 & 0.1249 & 0.1706 & $-$0.037 & 0.0986 & 0.0705 & 0.1065 & 0.1441 & $-$0.040 \\
        TuckER+EWC & 0.0679 & 0.0474 & 0.0732 & 0.0988 & \textbf{$-$0.014} & 0.0712 & 0.0433 & 0.0799 & 0.1160 & \textbf{$-$0.015} & 0.0598 & 0.0384 & 0.0671 & 0.0965 & \textbf{$-$0.017} \\
        \midrule
        MMTransE+FT & 0.0963 & 0.0330 & 0.1235 & 0.1922 & $-$0.070 & 0.1038 & 0.0310 & 0.1386 & 0.2320 & $-$0.076 & 0.0858 & 0.0271 & 0.1146 & 0.1899 & $-$0.083 \\
        MMDistMult+FT & 0.1175 & 0.0872 & 0.1228 & 0.1606 & \underline{$-$0.016} & 0.1247 & 0.0805 & 0.1357 & 0.1908 & \underline{$-$0.018} & 0.1064 & 0.0726 & 0.1158 & 0.1612 & \underline{$-$0.019} \\
        MMRotatE+FT & 0.1008 & 0.0777 & 0.1031 & 0.1331 & $-$0.038 & 0.1074 & 0.0720 & 0.1144 & 0.1588 & $-$0.041 & 0.0893 & 0.0633 & 0.0951 & 0.1307 & $-$0.045 \\
        \midrule
        MMTransE+EWC & 0.1014 & 0.0365 & 0.1316 & 0.1959 & $-$0.068 & 0.1089 & 0.0341 & 0.1472 & 0.2355 & $-$0.074 & 0.0912 & 0.0302 & 0.1232 & 0.1953 & $-$0.081 \\
        MMTransE+Replay & 0.1052 & 0.0373 & 0.1325 & \underline{0.2142} & $-$0.061 & 0.1125 & 0.0347 & 0.1475 & \underline{0.2564} & $-$0.066 & 0.0946 & 0.0309 & 0.1240 & \underline{0.2135} & $-$0.072 \\
        \midrule
        \textbf{MRCKG} & \textbf{0.2712} & \textbf{0.2186} & \textbf{0.2900} & \textbf{0.3578} & $-$0.065 & \textbf{0.2953} & \textbf{0.2254} & \textbf{0.3287} & \textbf{0.4367} & $-$0.070 & \textbf{0.2183} & \textbf{0.1578} & \textbf{0.2430} & \textbf{0.3206} & $-$0.076 \\
        \bottomrule
    \end{tabular}
\end{table*}
\begin{table*}[h]
    \centering
    \caption{Full results on MKG-Y benchmarks (Avg MRR / H@1 / H@3 / H@10 / BWT). Best \textbf{bold}, second best \underline{underlined}.}
    \label{tab:appendix-mkgy}
    \setlength{\tabcolsep}{1.8pt}
    \scriptsize
    \begin{tabular}{@{}l|ccccc|ccccc|ccccc@{}}
        \toprule
        Method          & \multicolumn{5}{c}{Entity} & \multicolumn{5}{c}{Higher} & \multicolumn{5}{c}{Equal} \\
        \cmidrule(lr){2-6} \cmidrule(lr){7-11} \cmidrule(lr){12-16}
                        & MRR & H@1 & H@3 & H@10 & BWT & MRR & H@1 & H@3 & H@10 & BWT & MRR & H@1 & H@3 & H@10 & BWT \\
        \midrule
        TransE+FT & 0.1243 & 0.0611 & 0.1403 & 0.2212 & $-$0.056 & 0.1587 & 0.0768 & 0.1811 & 0.2962 & $-$0.053 & 0.1462 & 0.0719 & 0.1668 & 0.2697 & $-$0.055 \\
        DistMult+FT & 0.1356 & 0.1073 & 0.1340 & 0.1961 & $-$0.050 & 0.1725 & 0.1344 & 0.1723 & 0.2617 & $-$0.047 & 0.1589 & 0.1258 & 0.1587 & 0.2382 & $-$0.049 \\
        ComplEx+FT & 0.1872 & 0.1685 & 0.1884 & 0.2192 & $-$0.021 & 0.2316 & 0.2084 & 0.2356 & 0.2845 & $-$0.020 & 0.2153 & 0.1938 & 0.2190 & 0.2614 & $-$0.021 \\
        TuckER+FT & 0.1835 & 0.1651 & 0.1799 & 0.2000 & $-$0.019 & 0.2274 & 0.2047 & 0.2253 & 0.2600 & $-$0.018 & 0.2108 & 0.1897 & 0.2089 & 0.2382 & $-$0.019 \\
        \midrule
        TransE+EWC & 0.1258 & 0.0689 & 0.1384 & 0.2158 & $-$0.055 & 0.1604 & 0.0865 & 0.1784 & 0.2886 & $-$0.051 & 0.1478 & 0.0810 & 0.1644 & 0.2628 & $-$0.053 \\
        TransE+Replay & 0.1315 & 0.0683 & 0.1503 & 0.2228 & $-$0.051 & 0.1672 & 0.0855 & 0.1932 & 0.2970 & $-$0.048 & 0.1543 & 0.0802 & 0.1783 & 0.2709 & $-$0.050 \\
        DistMult+EWC & 0.1389 & 0.1135 & 0.1419 & 0.1930 & $-$0.039 & 0.1758 & 0.1413 & 0.1815 & 0.2562 & $-$0.037 & 0.1624 & 0.1327 & 0.1677 & 0.2339 & $-$0.038 \\
        ComplEx+EWC & \underline{0.1907} & \underline{0.1716} & \underline{0.1937} & 0.2199 & $-$0.020 & \underline{0.2362} & \underline{0.2126} & \underline{0.2426} & 0.2856 & $-$0.019 & \underline{0.2197} & \underline{0.1977} & \underline{0.2256} & 0.2626 & $-$0.020 \\
        RotatE+EWC & 0.1412 & 0.1271 & 0.1389 & 0.1659 & $-$0.032 & 0.1793 & 0.1614 & 0.1783 & 0.2209 & $-$0.030 & 0.1659 & 0.1493 & 0.1650 & 0.2020 & $-$0.031 \\
        TuckER+EWC & 0.0914 & 0.0798 & 0.0934 & 0.1185 & \textbf{$-$0.013} & 0.1178 & 0.1012 & 0.1217 & 0.1602 & \textbf{$-$0.012} & 0.1087 & 0.0949 & 0.1123 & 0.1461 & \textbf{$-$0.013} \\
        \midrule
        MMTransE+FT & 0.1237 & 0.0531 & 0.1504 & 0.2201 & $-$0.065 & 0.1579 & 0.0667 & 0.1941 & 0.2946 & $-$0.061 & 0.1453 & 0.0623 & 0.1786 & 0.2680 & $-$0.064 \\
        MMDistMult+FT & 0.1498 & 0.1348 & 0.1485 & 0.1825 & \underline{$-$0.015} & 0.1896 & 0.1706 & 0.1900 & 0.2422 & \underline{$-$0.014} & 0.1752 & 0.1577 & 0.1756 & 0.2212 & \underline{$-$0.015} \\
        MMRotatE+FT & 0.1279 & 0.1151 & 0.1241 & 0.1505 & $-$0.035 & 0.1635 & 0.1472 & 0.1604 & 0.2018 & $-$0.033 & 0.1508 & 0.1357 & 0.1479 & 0.1840 & $-$0.034 \\
        \midrule
        MMTransE+EWC & 0.1302 & 0.0586 & 0.1603 & 0.2241 & $-$0.063 & 0.1657 & 0.0734 & 0.2062 & 0.2992 & $-$0.060 & 0.1529 & 0.0688 & 0.1903 & 0.2729 & $-$0.062 \\
        MMTransE+Replay & 0.1341 & 0.0595 & 0.1602 & \underline{0.2433} & $-$0.056 & 0.1704 & 0.0743 & 0.2057 & \underline{0.3243} & $-$0.053 & 0.1572 & 0.0697 & 0.1898 & \underline{0.2957} & $-$0.055 \\
        \midrule
        \textbf{MRCKG} & \textbf{0.2518} & \textbf{0.2195} & \textbf{0.2553} & \textbf{0.2973} & $-$0.060 & \textbf{0.3285} & \textbf{0.2791} & \textbf{0.3367} & \textbf{0.4053} & $-$0.056 & \textbf{0.3146} & \textbf{0.2748} & \textbf{0.3225} & \textbf{0.3815} & $-$0.058 \\
        \bottomrule
    \end{tabular}
\end{table*}

\subsection{Benchmark Construction}
\label{sec:appendix-dataset}

\subsubsection{Source Datasets}

Three publicly available MMKG datasets serve as the basis for our benchmarks:
\begin{itemize}[leftmargin=*,nosep]
    \item \textbf{DB15K}~\cite{liu2019mmkg}: derived from DBpedia, containing 12,842 entities, 279 relations, and approximately 99,028 triples. The graph is relatively dense, with 97.9\% visual coverage and 100\% textual coverage.
    \item \textbf{MKG-W}~\cite{xu2022relation}: derived from Wikidata, with 15,000 entities, 169 relations, and approximately 46,001 triples.
    \item \textbf{MKG-Y}~\cite{xu2022relation}: derived from YAGO, with 15,000 entities but only 28 relations and approximately 29,088 triples, making it a sparse and challenging graph.
\end{itemize}
For each dataset, we merge the original training, validation, and test triples into a single complete graph and re-partition it into an evolving sequence of $T{=}5$ snapshots using three strategies described below.

\subsubsection{Snapshot Split Strategies}

\textbf{Entity (progressive entity growth).}
All entities are sorted by degree in descending order. The initial snapshot $\mathcal{S}_0$ contains the top 35\% highest-degree entities together with their associated triples. The remaining entities are allocated to subsequent snapshots in decreasing proportions $[1, \frac{1}{2}, \frac{1}{3}, \frac{1}{4}]$. At each snapshot, only triples whose head and tail have both been introduced are included. This simulates a scenario where the core graph is established first, and peripheral entities join gradually, so the number of new triples decreases over time.

\textbf{Higher (BFS accelerated growth).}
Starting from the top 2\% highest-degree entities as seeds, breadth-first search is performed to obtain an entity visitation order. Target triple counts per step follow the ratio $[1 : 1.5 : 2 : 3 : 4]$, and entities are included along the BFS order accordingly. Later snapshots thus contain substantially more new knowledge, posing a greater forgetting challenge.

\textbf{Equal (BFS uniform growth).}
The top 1\% highest-degree entities serve as BFS seeds. Each step receives approximately the same number of triples ($\lfloor|\mathcal{T}|{/}T\rfloor$), with the final step absorbing all remaining triples. This provides a uniform incremental workload for evaluating model behavior under steady growth.

\subsubsection{Bridge Triples and Data Splitting}

To maintain structural connectivity across snapshots, starting from $\mathcal{S}_1$ each snapshot samples 15\% of the triples from the preceding snapshot's training set as \textbf{bridge triples} and merges them with the newly added triples. Because bridge triples involve previously seen entities and relations, they help the model preserve connections between old and new subgraphs during incremental training.

After merging, the triples at each step are randomly split into training, validation, and test sets at a $3{:}1{:}1$ ratio (random seed $42{+}i$, where $i$ is the snapshot index). Node degree, node betweenness centrality, and edge betweenness centrality are also precomputed for each training graph to support baselines such as IncDE.

\subsubsection{Multimodal Features}

Visual features are extracted using a frozen BEiT~\cite{bao2021beit} codebook encoder, yielding $m{=}8$ visual tokens per entity (codebook size 8192, raw dimension 32). Textual features come from a frozen BERT~\cite{devlin2018bert} word embedding layer, yielding $n{=}8$ text tokens per entity (vocabulary size 30522, raw dimension 768). All pretrained features remain frozen across snapshots and receive no gradient updates.

\subsection{Evaluation Metrics}
\label{sec:appendix-metrics}

\subsubsection{Filtered Ranking Protocol}

For each test triple $(h, r, t)$, a tail prediction query $(h, r, ?)$ is constructed. All candidate entities are scored and ranked. Before ranking, every known correct tail entity other than $t$ is removed from the candidate list (the \emph{filtered} setting), preventing valid triples from being counted as incorrect predictions. The resulting position of the correct answer is denoted $\mathrm{rank}(h,r,t)$.

\subsubsection{Link Prediction Metrics}

Let $\mathcal{Q}$ denote the test set containing $|\mathcal{Q}|$ query triples.

\textbf{Mean Reciprocal Rank (MRR):}
\begin{equation}
    \mathrm{MRR} = \frac{1}{|\mathcal{Q}|}\sum_{(h,r,t)\in\mathcal{Q}} \frac{1}{\mathrm{rank}(h,r,t)}
\end{equation}
MRR is the primary link prediction metric, as it is more sensitive to high-ranking predictions. It ranges in $(0, 1]$; higher is better.

\textbf{Hits@$K$ ($K \in \{1, 3, 10\}$):}
\begin{equation}
    \mathrm{Hits@}K = \frac{1}{|\mathcal{Q}|}\sum_{(h,r,t)\in\mathcal{Q}} \mathbb{1}[\mathrm{rank}(h,r,t) \leq K]
\end{equation}
Hits@$K$ measures the proportion of queries for which the correct answer appears within the top $K$ candidates. Hits@1 corresponds to exact match accuracy, and Hits@10 captures top-10 recall.

\textbf{Mean Rank (MR):}
\begin{equation}
    \mathrm{MR} = \frac{1}{|\mathcal{Q}|}\sum_{(h,r,t)\in\mathcal{Q}} \mathrm{rank}(h,r,t)
\end{equation}
MR is the arithmetic mean of all ranks; lower is better. Because it is sensitive to outlier ranks, it serves as a supplementary reference.

\subsubsection{Continual Learning Metrics}

\textbf{Average metrics.}
After training on all $T$ snapshots, the final model is evaluated on every seen test set $\mathcal{S}_0^{\text{test}}, \ldots, \mathcal{S}_{T-1}^{\text{test}}$, and the per-snapshot scores are averaged:
\begin{equation}
    \mathrm{Avg\;MRR} = \frac{1}{T}\sum_{i=0}^{T-1} \mathrm{MRR}(\mathcal{S}_i^{\text{test}})
\end{equation}
Avg Hits@$K$ and Avg MR are computed analogously. This metric reflects both the ability to acquire new knowledge and to retain old knowledge.

\textbf{Backward Transfer (BWT):}
\begin{equation}
    \mathrm{BWT} = \frac{1}{T-1}\sum_{i=0}^{T-2}\bigl(a_{T-1,i} - a_{i,i}\bigr)
\end{equation}
where $a_{j,i}$ is the MRR on $\mathcal{S}_i^{\text{test}}$ after completing training on $\mathcal{S}_j$. $\mathrm{BWT}{=}0$ indicates no forgetting, $\mathrm{BWT}{<}0$ indicates catastrophic forgetting (more negative means more severe), and $\mathrm{BWT}{>}0$ indicates positive backward transfer. In practice, BWT is typically negative; values closer to zero indicate less forgetting.

\subsubsection{Metrics.}
After training on each snapshot $\mathcal{S}_i$, MRR is evaluated on all seen test sets $\mathcal{S}_0^{\text{test}}$ through $\mathcal{S}_i^{\text{test}}$, from which we derive:
\begin{itemize}[leftmargin=*,nosep]
    \item $\mathrm{MRR_{new}}(i)$: MRR on the current snapshot $\mathcal{S}_i$ test set (plasticity).
    \item $\mathrm{MRR_{old}}(i)$: average MRR on all previous snapshots $\mathcal{S}_0,\dots,\mathcal{S}_{i-1}$ (stability).
    \item $\mathcal{S}_0$ retention: ratio of the final model's MRR on $\mathcal{S}_0^{\text{test}}$ to that obtained immediately after training on $\mathcal{S}_0$.
\end{itemize}


\subsubsection{Error Type Analysis}
\label{sec:app-case-study}

\textbf{Error classification criteria.}
On the final model trained on DB15K-Entity, filtered ranks are computed for all $\sum_{i=0}^{4}|\mathcal{S}_i^{\text{test}}|{=}7445$ test triples. Each prediction is classified according to the following rules:
\begin{enumerate}[leftmargin=*,nosep]
    \item \textbf{Correct} (Rank${=}$1): the model ranks the correct entity first.
    \item \textbf{Cold-start error}: the query involves an entity introduced in a later snapshot ($\mathcal{S}_2$ or beyond) and Rank${>}$10, indicating insufficient training for the new entity.
    \item \textbf{Forgetting error}: the query originates from an early snapshot ($\mathcal{S}_0$ or $\mathcal{S}_1$) and Rank${>}$10, indicating degradation of previously learned knowledge.
    \item \textbf{Cross-modal ambiguity}: Rank falls between 2 and 10, and the top candidate shares similar multimodal features with the correct entity (cosine similarity${>}$0.5).
    \item \textbf{Other error}: all remaining errors (Rank${>}$10) that do not fall into the cold-start or forgetting categories.
\end{enumerate}